\documentclass[letterpaper, 10 pt, conference]{ieeeconf}  

\IEEEoverridecommandlockouts                              

\usepackage{graphicx} 
\usepackage[position=bottom]{subfig}
\usepackage{multirow}
\usepackage{textcomp}
\usepackage{siunitx}
\usepackage{tabularx}
\usepackage{booktabs}
\usepackage{makecell}
\usepackage{hyperref}
\usepackage{cite}  
\usepackage{xcolor}

\usepackage{amsmath} 

\title{\LARGE \bf
DRAPER: Towards a Robust Robot Deployment and Reliable Evaluation for Quasi-Static Pick-and-Place Cloth-Shaping Neural Controllers}

\author{Halid Abdulrahim Kadi$^{1}$, Jose Alex Chandy$^{2}$, Luis Figueredo$^{2}$, Kasim Terzi{\'c}$^{1}$, Praminda Caleb-Solly$^{2}$ 
\thanks{$^{1}$ School of Computer Science, University of St Andrews
        {\tt\small }}%
\thanks{$^{2}$ School of Computer Science, University of Nottingham
        {\tt\small}}%
}

\begin{document}

\maketitle
\thispagestyle{empty}
\pagestyle{empty}


\begin{abstract}
Comparing robotic cloth-manipulation systems in a real-world setup is challenging. The fidelity gap between simulation-trained cloth neural controllers and real-world operation hinders the reliable deployment of these methods in physical trials. Inconsistent experimental setups and hardware limitations among different approaches obstruct objective evaluations. This study demonstrates a reliable real-world comparison of different simulation-trained neural controllers on both flattening and folding tasks with different types of fabrics varying in material, size, and colour. We introduce the DRAPER framework to enable this comprehensive study, which reliably reflects the true capabilities of these neural controllers. It specifically addresses real-world grasping errors, such as misgrasping and multilayer grasping, through real-world adaptations of the simulation environment to provide data trajectories that closely reflect real-world grasping scenarios. It also employs a special set of vision processing techniques to close the simulation-to-reality gap in the perception. Furthermore, it achieves robust grasping by adopting a tweezer-extended gripper and a grasping procedure. We demonstrate DRAPER's generalisability across different deep-learning methods and robotic platforms, offering valuable insights to the cloth manipulation research community. Please visit our project website \href{https://sites.google.com/view/draper-pnp}{https://sites.google.com/view/draper-pnp} for demonstration videos and code.
\end{abstract}

\begin{table*}[t!]
\caption{Real-World Flattening Performance of RGB-based Neural Controllers Deployed using DRAPER framework; NI (\%) represents normalised improvement, NC (\%)  represents normalised coverage, and SR represents flattening success rate; a trajectory is regarded as failure if it cannot reach above 95\% of NC within 20 steps. Human policy finishes execution in 10.8  $\pm$ 4.6 steps, diffusion policy 16.7 $\pm$ 4.5 steps, JA-TN 16.4  $\pm$ 5.0 steps and PlaNet-ClothPick 18.8 $\pm$ 3.1. We allow only one trial for each fabric where all corners are visible and other trials are randomly crumpled. JA-TN and diffusion policy outperform PlaNet-ClothPick substantially when trained against similar amount of data.}
\label{tab:flattening}
\centering
\resizebox{\textwidth}{!}{%
\begin{tabular}{c|ccc|ccc|ccc||ccc}
\hline
Method & \multicolumn{3}{c|}{PlaNet-ClothPick} & \multicolumn{3}{c|}{JA-TN} & \multicolumn{3}{c||}{Diffusion Policy} & \multicolumn{3}{c}{Human Policy} \\
\cline{1-13}
Fabric & NC & NI & SR & NC & NI & SR & NC & NI & SR & NC  & NI & SR \\
\hline
A & 45.6 $\pm$ 14.1 & 12.9 $\pm$ 13.5  & 0/5 & \textbf{71.9 $\pm$ 24.4} & \textbf{62.1 $\pm$ 32.2}  &  \textbf{2/5} & 68.5 $\pm$ 27.0 & 60.3 $\pm$ 32.4 & \textbf{2/5} & 92.1 $\pm$ 13.6 & 89.2 $\pm$ 18.7 & 4/5 \\

B & 75.6 $\pm$ 13.0  & 57.4 $\pm$ 17.2 & 1/5 &  \textbf{92.5 $\pm$ 15.9} & \textbf{88.7 $\pm$ 24.4} & \textbf{4/5}  &  85.2 $\pm$ 17.0 & 80.0 $\pm$ 22.9  & 3/5 &  99.0 $\pm$ 1.5 & 98.5 $\pm$ 2.1 & 5/5 \\

C & 90.0 $\pm$ 18.9 & 84.3 $\pm$ 29.1 & \textbf{3/5} & 81.7 $\pm$ 26.5 & 72.5$\pm$ 40.8 &  \textbf{3/5} & \textbf{96.6 $\pm$ 6.2} & \textbf{95.2 $\pm$ 7.8}  & \textbf{3/5} &  99.9 $\pm$ 0.2 & 99.8 $\pm$ 0.4 & 5/5 \\

D & 91.5 $\pm$ 15.6 &  87.2 $\pm$ 21.7  & \textbf{4/5}  & 88.0 $\pm$ 25.2   & 82.5 $\pm$ 36.1 & 3/5 & \textbf{96.9 $\pm$ 4.0}   & \textbf{95.7 $\pm$ 5.4}  & \textbf{4/5} & 99.2 $\pm$ 0.5 & 98.9 $\pm$  0.7 & 5/5 \\

E &  68.1 $\pm$ 25.1 & 54.3 $\pm$ 36.1 & 1/5 &  \textbf{94.1 $\pm$  12.9} & \textbf{91.1 $\pm$ 18.9} & \textbf{4/5} & 87.8 $\pm$ 15.2 &  82.6 $\pm$ 20.5  & 3/5  & 99.4 $\pm$ 1.3 & 98.7 $\pm$ 2.47 & 5/5\\

F &  	72.4  $\pm$ 12.8 & 58.5 $\pm$ 12.6  &  0/5 & \textbf{98.29  $\pm$ 2.79} & \textbf{97.8  $\pm$  3.7} & \textbf{4/5}  & 81.7 $\pm$ 25.0 & 77.2  $\pm$ 31.5  & 3/5  & 99.3 $\pm$ 1.1  & 98.7 $\pm$ 2.5 & 5/5\\

\hline
Real Total &  73.8 $\pm$ 22.0& 59.1 $\pm$ 32.7 &  9/30 &  \textbf{87.8  $\pm$  20.1} & \textbf{82.5 $\pm$ 28.9}  & \textbf{20/30} & 86.1 $\pm$ 19.1  &  81.8 $\pm$ 24.0 & 18/30  & 98.1 $\pm$ 5.8 & 97.3 $\pm$ 8.0 & 29/30 \\

\end{tabular}%
}

\end{table*}

\begin{table*}[t!]
\caption{Real-World Folding Performance of RGB-based Neural Controllers Deployed using DRAPER framework; IoU (\%) indicates the maximum intersection-over-union between the cloth and the goal mask, while SR reflects the folding success rate based on human judgment. Both JA-TN and diffusion policies can be successfully deployed in real-world using DRAPER, with the latter showing a more robust capability that approaches human performance.  } 
\label{tab:folding}
\centering
\resizebox{\textwidth}{!}{%
\begin{tabular}{l||cc|cc||cc|cc||cc|cc||cc}
\hline
Folding Tasks & \multicolumn{4}{c||}{All Corner Inward} & \multicolumn{4}{c||}{Corners Edge Inward} & \multicolumn{4}{c||}{Digonal Cross} & \multicolumn{2}{c}{Total} \\
\cline{1-15}
Fabric & \multicolumn{2}{c|}{A} & \multicolumn{2}{c||}{ B} & \multicolumn{2}{c|}{A} & \multicolumn{2}{c||}{B} & \multicolumn{2}{c|}{A} & \multicolumn{2}{c||}{B} & \multicolumn{2}{c}{all} \\
\cline{1-15}
 Method \textbackslash Metrics & IoU $\uparrow$ & SR $\uparrow$ & IoU $\uparrow$ & SR $\uparrow$ & IoU $\uparrow$ & SR $\uparrow$ & IoU  $\uparrow$& SR $\uparrow$ & IoU $\uparrow$ & SR $\uparrow$ & IoU $\uparrow$ & SR $\uparrow$ & IoU $\uparrow$ & SR $\uparrow$\\
\hline
Foldsformer & 83.2 $\pm$ 16.6 & \textbf{4/5}   & \textbf{90.2 $\pm$  5.9}  & 4/5  & 83.0 $\pm$ 5.7   &  0/5 & 77.2 $\pm$ 3.3 & 0/5  & 87.1 $\pm$ 1.6 & \textbf{5/5} & \textbf{89.3 $\pm$ 1.3}  & \textbf{5/5}  & 85.0 $\pm$ 8.4  & 18/30  \\
JA-TN & 83.7 $\pm$ 7.8  & 3/5 & 84.1 $\pm$ 8.6 &  3/5  &  84.2  $\pm$ 7.4 & \textbf{3/5}   & 79.8  $\pm$  10.5  & 1/5 & 85.6 $\pm$ 9.1  & 4/5 &   83.8 $\pm$ 10.1 & 4/5  & 83.5 $\pm$ 8.4  & 18/30  \\
Diffusion Policy & \textbf{91.0 $\pm$ 2.1}  & \textbf{4/5} & 90.0 $\pm$ 1.3  & \textbf{5/5} & \textbf{89.0 $\pm$ 4.8}  & \textbf{3/5} & \textbf{84.6 $\pm$ 9.8} & \textbf{3/5}  &  \textbf{89.1 $\pm$ 5.5}& \textbf{5/5} & 85.5 $\pm$ 3.9 & \textbf{5/5} &  \textbf{88.2 $\pm$ 5.4} & \textbf{25/30}\\
\hline
Human Policy & 93.3 $\pm$ 1.9 & 5/5 & 91.4 $\pm$ 3.6 & 5/5 & 85.5 $\pm$ 11.1 & 4/5 & 90.1 $\pm$ 2.9 & 5/5 & 87.0 $\pm$ 1.5 & 5/5 & 85.8 $\pm$ 3.0 & 4/5 & 88.9 $\pm$ 5.6 & 28/30\\

\end{tabular}%
}

\end{table*}

\section{Introduction}
Cloth-shaping controllers are difficult to compare
in real-world experiments. There has been impressive progress in recent years on developing vision-based data-driven controllers for cloth flattening and folding, including both reinforcement learning \cite{hoque2022visuospatial, yan2021learning, lin2022learning, lee2021learning, kadi2024planet} and imitation learning \cite{seita2020deep, weng2022fabricflownet} methods. Whilst many studies have compared these approaches in simulated environments \cite{kadi2024mjtn, lin2020softgym, mo2022foldsformer} and reported successful transfer of simulation-trained policies to the real world individually \cite{kadi2023data}, we are not aware of any comparative study of different algorithms on a real robot or any analysis of how they generalise across different cloth materials, grippers, or robot platforms. The challenges of replicating real-world experiments in this domain is mainly due to the nuance difference between the vision processing strategies and various real-world grasping errors, such as misgrasping, cloth dropping, and grasping multiple layers \cite{hoque2022visuospatial, weng2022fabricflownet, mo2022foldsformer}. In this paper, we analyse the main obstacles to real-world deployment in the cloth-shaping domain and present an overarching real-world comparison of four recent cloth-shaping algorithms on a variety of flattening and folding tasks, across different robot platforms and fabric types.

\begin{figure}[t]
    \centering
    \includegraphics[width=0.5\textwidth]{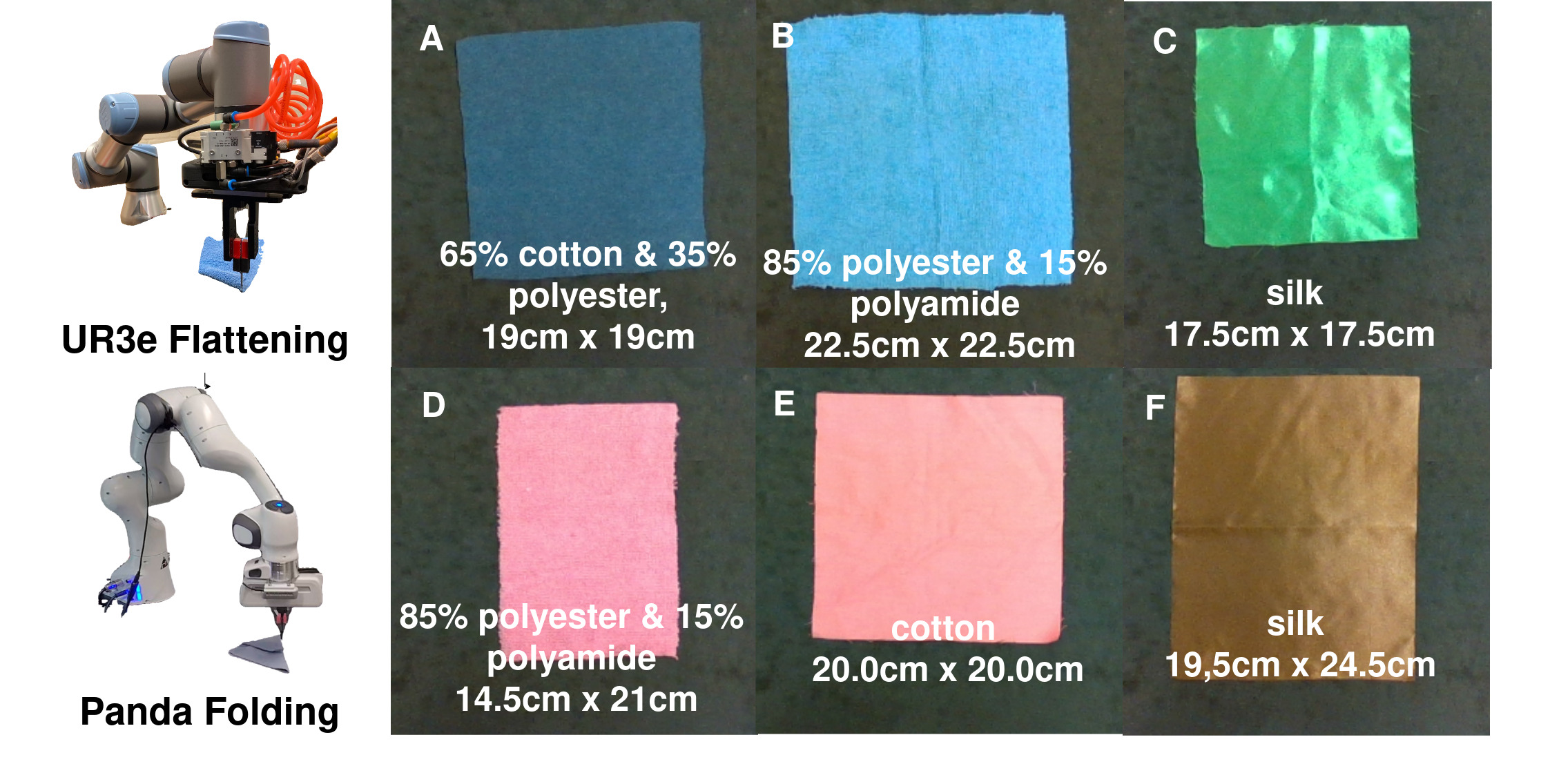}
    \caption{Robust Folding and Flattening Deployment of Neural Controllers on Various Robot Platforms. We examine the performance of these controllers on 6 different towels, varying in terms of material, colour and length-to-width ratios.} 
    \label{fig:framework}
\end{figure}




One of the primary challenges in deploying cloth manipulation systems in real-world settings can be attributed to the difference between simulated and physical environments, particularly in the behaviour of the fabric itself and the interaction between the end-effector and the fabric. Two main strategies to address these challenges are: (1) enhancing physics engines to more accurately model cloth behaviour to close the reality-to-simulation gap which can be very difficult to achieve; and (2) developing algorithms that learn directly in the real world, which can be limited to specific use cases and by the scarcity of available demonstrations \cite{lee2021learning}. 

Following the former strategy, we choose to redesign the real-world deployment framework for simulation-trained cloth-shaping algorithms by integrating grasping uncertainty into simulation environments to create more robust policies, enhancing the reliability of real-world pick action. This results in our proposed framework for \textbf{D}eep \textbf{R}obotic \textbf{A}daptive \textbf{P}ick-and-plac\textbf{E} controllers for \textbf{R}eal-world cloth manipulation (DRAPER). In general, we make the following contributions: 

1.  We reliably compare and contrast 4 different simulation-trained neural policies in multiple real-world robotic setups on 6 different types of fabrics that varies in the length-width ratio and material properties as shown in Table \ref{tab:flattening} and \ref{tab:folding}. To our knowledge, this is the first such real-world comparison of garment manipulation algorithms. We identify difficult cases whilst manipulating really soft fabrics, such as the ones made from a mixture of cotton and polyester, as different parts of the fabrics tend to stick together compared to other materials.

2. We achieve this through proposed DRAPER deployment framework (see Figure \ref{fig:framework}). It adopts real-world adaptation in simulation for data-collection, applies a standardised vision processing algorithm for training and real-image processing based on JA-TN \cite{kadi2024mjtn}, and employs tweezer extension to improve grasping precision as in Hoque et al. \cite{hoque2021lazydagger}. To reduce misgrasping scenarios, the proposed framework also includes a grasping procedure for improving grasping reliability.

3. We successfully adapt, for the first time, the diffusion policy \cite{chi2023diffusionpolicy} in the pick-and-place domain and show that it achieves the best real-world and simulation results for both towel-flattening and towel-folding tasks. 

4. We conduct extensive experiments in numerous settings to showcase the DRAPER framework's compatibility with Franka Emika Panda and Universal Robots UR3e in both "eye-in-hand" and "hand-to-eye" configurations under both ROS1 Noetic and ROS2 Humble settings; this demonstrates that our approach works across different robot arms, hardware, and software.

\begin{figure*}[ht!]
    \centering
    \includegraphics[width=1.0\textwidth]{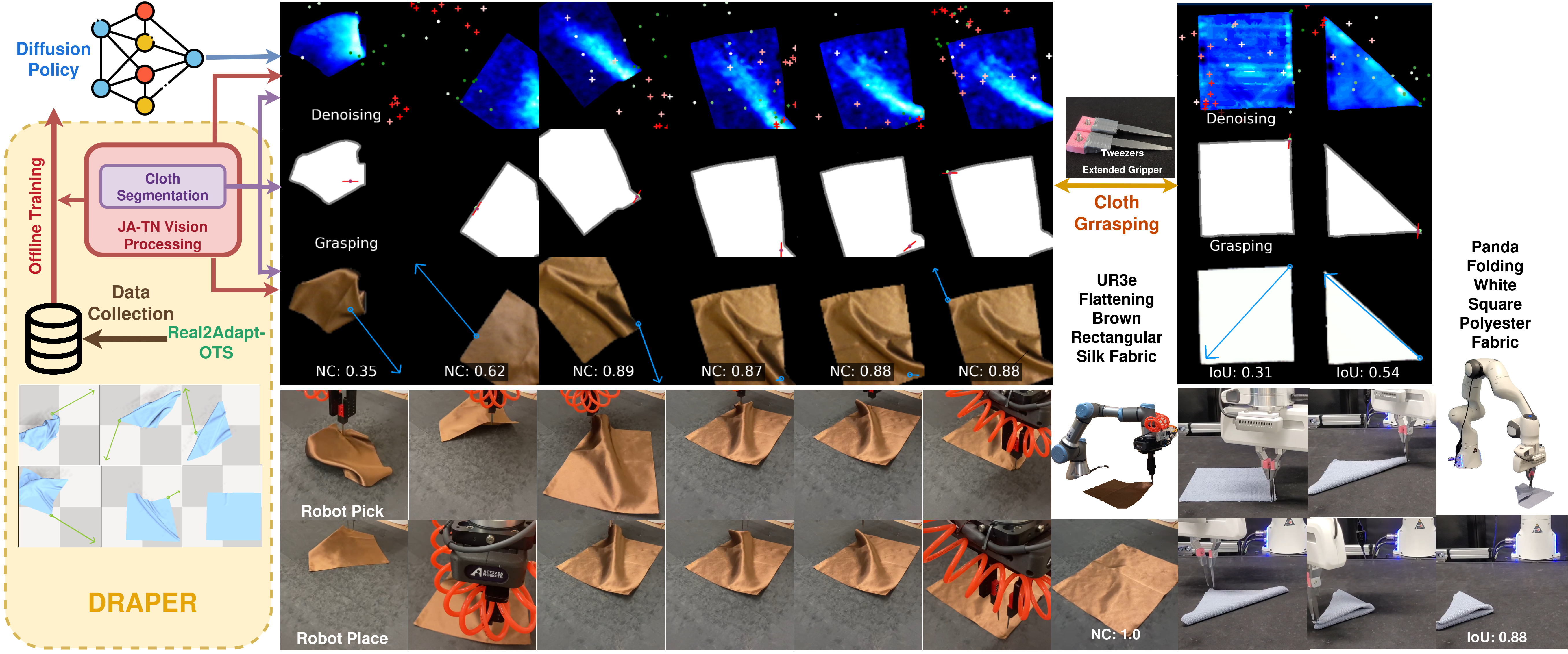}
   
    \caption{DRAPER framework that deploys RGB-D Diffusion Policy to Real-Robot Setups. Our real-world adaptation of Oracle Towel Smoothing (RealAdapt-OTS) policy is shown as green arrows on the trajectory generated by the proposed \textit{RealAdapt Towels} simulation environment in the yellow box. The first and third row trajectories from the top show the vision input and actions (blue arrows) induced by the diffusion policy. In the top row, diffusion-generated pick positions are marked by green dots and place positions by red crosses with lighter colours indicating earlier denoising steps. Purple dots in the ``Graspin'' images at the second row represent readjusted pick positions on eroded mask images, and the red line indicates gripper orientation. The neural controller preprocesses the RGB (3rd row) and depth images (1st row) with the JA-TN vision processing strategy (red box) \cite{kadi2024mjtn}. Performance is evaluated using Normalised Coverage (NC) for flattening and Intersection-over-Union (IoU) for folding. The last two trajectories demonstrate synchronised pick-and-place manipulation by UR3e and Panda robot arms for flattening and folding individually. Neural controllers can learn from the offline data collected by RealAdapt-OTS policy in our improved \textit{RealAdapt Towels} simulated benchmark environment for robust deployment to reality using our grasping protocol.} 
    \label{fig:framework}
\end{figure*}

\section{Related Work}
Numerous methods have been applied to single-gripper quasi-static pick-and-place (QSP\&P) cloth-shaping problems but have not been tested in the same robotic environment. We provide such a comparison in this work. 

\subsection{Neural Controllers for Cloth Shaping}
Several influential learning-based cloth-shaping controllers are based on model predictive control (MPC) including mesh-based dynamic planning method VCD \cite{lin2022learning} and vision-based planning method VSF \cite{hoque2022visuospatial}. PlaNet-ClothPick \cite{kadi2024planet}, on the other hand, uses a Recurrent State Space Model with variational inference and a domain-specific planning algorithm for fabric flattening; it samples pick locations on the cloth mask using a learned latent dynamic model and optimises policies based on predicted rewards. Among these MPC-based methods, we chose PlaNet-ClothPick as a representative to demonstrate the effectiveness of our deployment approach.

Foldsformer \cite{mo2022foldsformer}, which is a variant of TimeSformer \cite{bertasius2021space},  utilises a space-time attention mechanism to capture the instruction feature of folding demonstrations. Another goal-directed method, FabricFlowNet \cite{weng2022fabricflownet}, achieves multi-step fabric folding from canonical states using optical flow; however, its performance is inferior to Foldsformer \cite{mo2022foldsformer} which we adopt as a SoTA representative for transformer-based controllers to benchmark folding tasks. We also evaluate the off-the-shelf FoldsFormer folding policy \cite{mo2022foldsformer} with our DRAPER deployment framework.

JA-TN \cite{kadi2024mjtn} is a towel-shaping controller that extends the TransporterNet architecture \cite{zeng2021transporter} with joint-probability action inference for improving multi-modality. It uses convolutional neural networks to directly predict pick and place action heatmaps from images. 
It has demonstrated different types of folding from flattened and crumpled states in simulated environments but has not been extensively evaluated in a real-world experiment. We test JA-TN in the real world as a representative of popular behaviour cloning controllers with state-action value maps as outputs.

Chi et al.\ \cite{chi2023diffusionpolicy} apply diffusion models \cite{ho2020denoising} to behavioural cloning for effective visuomotor control policies; it has been examined successfully on wide range of robot control tasks in Robomimic \cite{mandlekar2021matters} and Franka Kitchen \cite{gupta2019relay} as well as in bimanual shirt folding, where it is trained on 300 demonstrations through a VR setup, achieves a 15/20 success rate with goal-conditioned low-level position control. Our research introduces the first application of the diffusion policy in QSP\&P cloth-shaping, further demonstrating the reliability of our real-world DRAPER deployment framework. In our framework, we leverage oracle policies in simulation environments to collect demonstrations, eliminating the need for costly VR human demonstrations.

\subsection{Vision Processing for Real-World Deployment} \label{sec:vision-process} Visual processing of cloth images often relies on cloth segmentation, with colour thresholding being a common heuristic.
VCD \cite{lin2022learning} focuses on extracting the point cloud of the cloth using segmentation in the target region. Several methods, including FabricFlowNet \cite{weng2022fabricflownet} and Foldsformer \cite{mo2022foldsformer}, utilise cloth-masked depth images as input to their models --- they subtract the difference between the average height of the real and the simulated camera to align depth between real and simulated images. In contrast, VSF \cite{hoque2022visuospatial} employs a domain-randomised \cite{tobin2017domain} strategy centred around the target fabric colour without relying on cloth segmentation. JA-TN \cite{kadi2024mjtn} also adopts domain randomisation in simulation but allows for a wider range of fabric colours. It adds Gaussian noise and blurring to the simulated depth image during training. In addition, it rescales and flips the values of the both real and simulated depth image between 0 and 1 in learning and real-world inference. Furthermore, it employs a ``segment-anything'' model \cite{kirillov2023segany}, combined with a heuristic mask selection, to mask the real-world images. We base DRAPER's vision processing on JA-TN.

\subsection{Robot Setup for Cloth-Shaping Controllers} Franka Emika Panda \cite{zhang2020modular} with its original gripper has been extensively tested on real-world QSP\&P cloth-shaping tasks \cite{lin2022learning, lee2021learning, weng2022fabricflownet, mo2022foldsformer}. VSF is a notable exception in using a Da Vinci robot with its pincer-like end-effector \cite{hoque2022visuospatial}, while LazyDagger \cite{hoque2021lazydagger} adopts a tweezer extension on the end-effector of a ABB YuMi
robot. JA-TN \cite{kadi2024mjtn} uses a stick-like 3D-printed extension on a UR3e robot, but only qualitative results are shown. This limited diversity in robotic setup presents a significant generalisability gap in the field, which our study addresses by demonstrating skill transferability of the examined method deployed by our DRAPER framework to different robotic platforms.

The materials used in the literature are often made of cotton, silk, polyester and polyamide materials with side lengths from 7cm to 45cm, typically using a single cloth per experiment. Our research extends this by incorporating multiple fabric types within the same experiment, addressing another crucial gap in existing studies \cite{lin2022learning, kadi2024mjtn}. 

In addition, the community has explored various camera options to capture fabric features, including Zivid One Plus \cite{hoque2022visuospatial}, Azure Kinect camera \cite{lin2022learning}, and most commonly, Intel RealSense D435 \cite{weng2022fabricflownet, lee2021learning, mo2022foldsformer, kadi2024mjtn}. While VCD adopts a side-view camera, other methods typically employ top-down settings, where the camera is mounted between 0.7 to 0.9 meters above a foamy table surface. Lee et al.\   \cite{lee2021learning}, FabricFlowNet and Foldsformer adopt ``eye-in-hand'' coordination between the end-effectors and the camera; others like VCD and VSF employ ``hand-to-eye'' configuration only. Our research bridges this gap by demonstrating the effectiveness of a single framework across multiple camera configurations, further demonstrating the versatility and applicability of DRAPER. 

\subsection{Real-World Performance of Cloth-Shaping Controllers} 
Due to these differences, to our knowledge, no comprehensive study exists to examine the real-world performance of the cloth-shaping neural controllers. Nevetherless, many individual studies reported their own robotic examinations. The transferred algorithm of VCD achieves 94\% coverage for cotton square fabric and 95\% for silk square fabric \cite{lin2022learning}. JA-TN reported a 7/10 success rate for a square and 10/10 for a small rectangular synthetic fabric \cite{kadi2024mjtn} but with human-in-the-loop pick-and-place execution where the policy is generated by the network through real camera input. Lee et al.\ \cite{lee2021learning} examine six folding tasks, achieving perfect success rates for four easier folds but struggling with more complex ones. FabricFlowNet and Foldsformer outperform Lee et al.\ on comparable folding tasks \cite{weng2022fabricflownet, mo2022foldsformer}. Notably, FabricFlowNet reaches 89\% IoU for one-step folding and 60\% IoU for multi-step folding, while Foldsformer achieves around 90\% IoU for various foldings and outperforms FabricFlowNet in corresponding tasks. Whilst various methods can be compared in the SoftGym simulation benchmark environment \cite{lin2020softgym}, proper real-world comparisons are lacking; typical reported sim2real system failures stem from misgrasping and grasping multiple layers \cite{mo2022foldsformer, weng2022fabricflownet, lin2022learning, kadi2024mjtn}. By allowing various methods to be more easily deployed in a real-world experiment across different platforms, materials, and camera configurations, we present a comprehensive real-world comparison of multiple state-of-the-art methods.


\section{DRAPER framework}

We train and deploy neural controllers such as PlaNet-ClothPick \cite{kadi2024planet}, JA-TN \cite{kadi2024mjtn}, as well as a general visuomotor diffusion policy \cite{chi2023diffusionpolicy} (Section \ref{sec:diffusion-policy}) with our DRAPER framework (see Fig.\ \ref{fig:framework}). It stands for \textbf{D}eep \textbf{R}obotic \textbf{A}daptive \textbf{P}ick-and-plac\textbf{E} controllers for \textbf{R}eal-world cloth manipulation. It incorporates real-world grasping scenarios in simulation and improves the Oracle policies based on the real-world adapted dynamics for providing expert demonstrations that are more closely reflect the real-world grasping scenarios (Section \ref{sec:real-world-adapt}). It also adopts JA-TN's vision processing for training  and real-image processing \cite{kadi2024mjtn} (Section \ref{sec:vision-process}). In addition, it employs a tweezer-extension on the end-effector of a robot arm to improve grasping precision on fabrics as shown by Hoque et al. \cite{hoque2021lazydagger}. Furthermore, it also includes a grasping procedure for
improving grasping reliability and reducing misgrasping scenarios (Section \ref{sec:cloth-grasp}).

\subsection{Real-World Adapted Simulation and Oracle Policy}  \label{sec:real-world-adapt}
For resulting in a simulation-trained policy that can be deployed to real-world pick-and-place towel manipulation, we must create an oracle policy specifically designed to address real-world problems. This involves considering three key aspects: 1) modelling realistic colours, including the same colour on both sides, 2) avoiding gripping challenging points with multiple layers, and 3) enabling the policy using parallel grippers to expose hidden corners. We extend the existing oracle policy OTS from JA-TN \cite{kadi2024mjtn} with these enhancements.

Grasping multiple layers of an article is a common phenomenon in real-world cloth manipulation. It is exacerbated by the fact that most simulation-trained neural controllers do not consider this factor during learning. We propose a new benchmark environment \textit{RealAdapt Towels} that incorporates real-world behaviours of the cloth-gripper interaction along with a corresponding improved oracle flattening policy to enhance the performance of the transferred deep policies. We enhance the \textit{Rainbow Rectangular Fabrics} environment \cite{kadi2024mjtn} by making the fabric share the same colours on both sides, including misgrasping chances and modifying the gripper behaviour to grasp a cluster of particles around the picker instead of precisely picking the closest cloth particle.

The quality of a behaviour cloning controller in the real-world application depends significantly on the quality of the expert demonstrated trajectories. The Oracle Towel Smoothing (OTS) policy employed by JA-TN \cite{kadi2024mjtn} leads to many failing scenarios in real-world deployment. As it typically chooses to unfold a fold on the top of the cloth and reveal a corner if it is hidden underneath the fabric, this policy only works well if a gripper can pick precisely the target particle of the cloth without grasping multiple layers; this rarely occurs in a real-world setup with a naive parallel gripper.

We extend the Oracle policy to unfold the fabric by first dragging the target corner area to the far end of the unfolding direction, then dragging the opposite corner to complete the unfolding. It also reveals hidden corners by dragging parts of the towel located on the opposite side. Whilst preserving many original characteristics of OTS, we rectify another significant oversight to ensure the entire fabric remains within the camera's field of view after flattening, preventing scenarios where portions of the smoothed cloth become obscured. We call the new policy RealAdapt Oracle Towel Smoothing (RealAdapt-OTS).


\subsection{Cloth Grasping} \label{sec:cloth-grasp}  


The original Franka Emika Panda gripper struggles with fabric attachment and precise grasping, and lacks precision for tasks involving edges and corners (see Table \ref{tab:cloth_fold_analysis}). We adopt a simple tweezers-extended gripper \cite{hoque2021lazydagger} (Figure \ref{fig:framework}) integrated into the 3D-printed base, significantly improving control and accuracy. The new gripper may still encounter misgrasping issues due to the precision of action-space conversion. To mitigate this, we implement two key strategies: (1) positioning the pick position slightly inside the cloth-mask border and (2) aligning the picking angle parallel to the cut-line of the pick-point when on a cloth edge. Notably, we refrain from incorporating depth estimation methods, allowing the gripper to descend as deeply as possible. 


\section{Adaptation of Diffusion Policy in Quasi-Static Pick-and-Place} \label{sec:diffusion-policy}

To demonstrate the robustness and versatility of our method, we apply our framework to diffusion policy \cite{chi2023diffusionpolicy} that has never before been used in the QSP\&P domain. This novel application serves to validate the reliability and adaptability of our framework DRAPER across different algorithmic approaches while exploring the potential for expanding the toolset available for ${\text{QS}}$P\&P tasks in fabric manipulation.

Denoising Diffusion Probabilistic Models (DDPMs) \cite{sohl2015deep, ho2020denoising} are a type of generative model that has gained popularity in the image and video generation \cite{ramesh2022hierarchical} and various control tasks. The diffusion process involves two key steps: a forward diffusion process and a learned reverse denoising process using a noise predictor. Chi et al.\ \cite{chi2023diffusionpolicy} adopt DDPMs to approximate the conditional policy distribution $\pi(\boldsymbol{\tau}^{\mathcal{A}}|\boldsymbol{\tau}^{\mathcal{X}})$, where $\boldsymbol{\tau}^{\mathcal{A}} \equiv \{\boldsymbol{a}_i\}_{i=t}^{t+T_p}$ is a sequence of action for prediction horizon $T_p$ from an arbitrary step $t$; and, $\boldsymbol{\tau}^{\mathcal{X}} \equiv \{\boldsymbol{x}_i\}_{i=t}^{i=T_{h}}$ is a sequence of history observation with horizon $T_h$ ($T_h \leq T_p$). Inference denoises the policies for the prediction horizon $T_p$ and only uses $T_a$-step actions on the environment. A vision encoder produces the vision features $\boldsymbol{e}_x$ by removing the final fully connected layer. The method can also have an internal state vector $s$ that is concatenated with the $\boldsymbol{e}_x$ to produce the conditional global feature $\boldsymbol{e}_g$.

The DDMP noise scheduler uses the following formula to add noise to the actions
\begin{equation}
    \boldsymbol{\tau}^{\mathcal{A}}_{k} \sim \mathcal{N}(\sqrt{1-\beta_k}\boldsymbol{\tau}^{\mathcal{A}}_{k-1}, \beta_k \boldsymbol{I}).
\end{equation}
Expanding the above distribution, we can directly get the ground truth noisy actions from $A_{0}$ for denoising step $k$, i.e., 
$
    \boldsymbol{\tau}^{\mathcal{A}}_{k} \sim \mathcal{N}( \overline{\alpha}_k  \boldsymbol{\tau}^{\mathcal{A}}_{k-1}, \sqrt{(1-\overline{\alpha}_k)} I)$
where  $\overline{\alpha}_k = \prod_{t=0}^k (1-\beta_t)$. 
Whilst in the reverse denoising, the sampled noisy actions $\boldsymbol{\tau}^{\mathcal{A}}_{K} \sim \mathcal{N}(\boldsymbol{0}, \boldsymbol{I})$ go through $K$ iterations of diffusion process, where a diffusion step $k$, last diffused policies $A_{k+1}$ and the global condition $z$ are fed into noise prediction network $\epsilon_\theta$ to produce prediction noises for step $k$ that is used to denoise from $\boldsymbol{\tau}^{\mathcal{A}}_{k}$: 
\begin{equation}
    \boldsymbol{\tau}^{\mathcal{A}}_{k-1} = \frac{1}{\sqrt{1-\beta_k}}\Big(\boldsymbol{\tau}^{\mathcal{A}}_{k} - \frac{\beta_k}{\sqrt{1-\overline{\alpha}_k}} \epsilon_{\theta}(\boldsymbol{\tau}^{\mathcal{A}}_{k}, k, \boldsymbol{e}_g)\Big) + \sigma_k \boldsymbol{\epsilon},    
\end{equation}
where $\epsilon \sim \mathcal{N}(0, I)$ and $\sigma^2_k = \tilde{\beta}_k = \frac{1-\overline{\alpha}_{k-1}}{1-\overline{\alpha}_k} \beta_{k}$ \cite{ho2020denoising}; square cosine scheduler for $\beta_k$ proposed in iDDPM \cite{nichol2021improved} is used for better performance.

By maximising the log-likelihood of the conditional policy distribution, the evidence lower bound loss function becomes:
\begin{equation}
    \mathcal{L} = MSE\Big(\epsilon_k,  \epsilon_\theta\big( \sqrt{\overline{\alpha}_k}\boldsymbol{\tau}^{\mathcal{A}}_{0} + \sqrt{(1-\overline{\alpha}_k})\epsilon_k , k, \boldsymbol{e}_g\big)\Big), \label{eqn:diffusion-policy-loss}
\end{equation}
where $\boldsymbol{\epsilon}_k \sim \mathcal{N}(0, I)$. The method uses a Gaussian distribution to normalise the input images and get the statistics from the dataset during the training.

In order to apply the method in folding tasks, we disable the down and up sampling of its 1-D U-Net \cite{ronneberger2015u} noise predictor for $T_h$ and $T_p$ as 1, but we keep it enabled for flattening with $T_h$ as 2 and $T_p$ as 4. We also discard the internal state vector $s$ for both flattening and folding domains. We train the policy with standard centre-pivot rotation and flipping augmentation along with vision processing strategy proposed by JA-TN \cite{kadi2024mjtn}.

\section{Experiments}

We conduct shaping experiments on a dark-grey foamy surface with Realsense D435i cameras --- UR3e in "hand-to-eye" configuration (0.74 m above the surface) and Panda in "eye-in-hand" configuration (0.57 m above the surface on taking images). We use the MoveIt! library \cite{moveit2documentation} to execute QSP\&P trajectory, as is common in this domain \cite{weng2022fabricflownet, mo2022foldsformer}. During physical trials, humans must reset the initial states for towel-shaping tasks, especially folding, and they also need to manually set the goal states for each trial for setup evaluation \cite{weng2022fabricflownet, mo2022foldsformer, lee2021learning, lin2022learning}. We examine the deployment of flattening controllers on 6 different fabrics varied in terms of colour, size, material and shapes (Figure \ref{fig:framework}). We tested folding performance on Fabrics A and B which are most suitable for folding due to their softness; each fabric was tested 5 times consecutively for each shaping task.

\begin{table*}[t!]
\caption{DRAPER Ablation Study using UR3e Robot for Flattening Rectangular Polyester Towel (Fabric D) with Diffusion Policy. NC (\%) represents normalised coverage, NI (\%) represents normalised improvement, SR represents flattening success rate, and S2F represents number of steps until finish/success; a trajectory is regarded as a failure if it cannot reach above 95\% of NC within 20 steps. All reported statistics are collected from real-world experiments. The ablation in the left columns uses the same RGB-D diffusion controller but with different grasping ablations; they evaluate on \textbf{pick position (Pos)} and \textbf{orientation (Orien) adjustment (Adj)}. The evaluation in the right columns uses the default grasping protocol of DRAPER but with controllers using different input types, image processing and trajectory data: (1) \textbf{Depth-Only} and \textbf{RGB-Only} ablations use the default image processing of DRAPER, but the two policies are trained using depth-only and RGB-only inputs, respectively. (2) \textbf{OTS in RRN} represents the RGB-D policy that trains on the dataset collected by Oracle Towel Smoothing (OTS) policy in \textit{Rainbow Rectangular Fabrics (RRN)} environments. (3) \textbf{Naive Depth} represents the RGB-D policy that is trained without further depth processing apart from Gaussian noise, and deployed by adding the difference between the simulation depth and the real-world depth. All aspects of the DRAPER framework is significant for the reliable deployment of a neural controllers.}
\label{tab:ablation-real}
\centering
\resizebox{\textwidth}{!}{%
\begin{tabular}{ccccc|cccccc}
\hline
Ablations  & NC $\uparrow$ & NI $\uparrow$ & S2F $\downarrow$ & SR  $\uparrow$ & Ablations (continue) & NC  $\uparrow$ & NI  $\uparrow$ & S2F $\downarrow$ & SR  $\uparrow$  \\
\hline
Default  & 	\textbf{89.6 $\pm$ 	17.2} & \textbf{83.5 $\pm$ 26.8} & \textbf{14.7 $\pm$ 5.5}  & \textbf{8/10} & Depth-Only & 86.5 $\pm$ 18.6 & 77.6 $\pm$ 30.2 & 16.9 $\pm$ 5.0 & 6/10\\
No Pos\&Orien Adj & 85.4 $\pm$ 16.6 & 67.6 $\pm$ 34.2 & 18.4 $\pm$ 2.5 & 5/10 & Naive Depth & 84.5 $\pm$ 16.3 & 74.2 $\pm$ 27.5 & 17.7 $\pm$ 3.9 & 4/10 \\
No Pos Adj & 87.4 $\pm$  13.7 & 79.9 $\pm$ 22.2 & 	18.2 $\pm$ 4.0 & 4/10 & OTS in RRN & 80.0 $\pm$ 20.4 & 66.9 $\pm$ 32.7 & 18.3 $\pm$ 3.6 & 4/10
\\ 
No Orien Adj & 76.9 $\pm$ 17.8 & 64.0 $\pm$ 28.2 & 18.9 $\pm$ 3.4 & 2/10 & RGB-Only & 87.5 $\pm$ 12.7 & 79.5 $\pm$ 20.5 & 	18.3 $\pm$ 3.9 & 5/10
\end{tabular}%
}
\end{table*}

\begin{figure*}[t!]
    \centering
    \subfloat[Normalised Coverage]{{\includegraphics[width=0.33\textwidth]{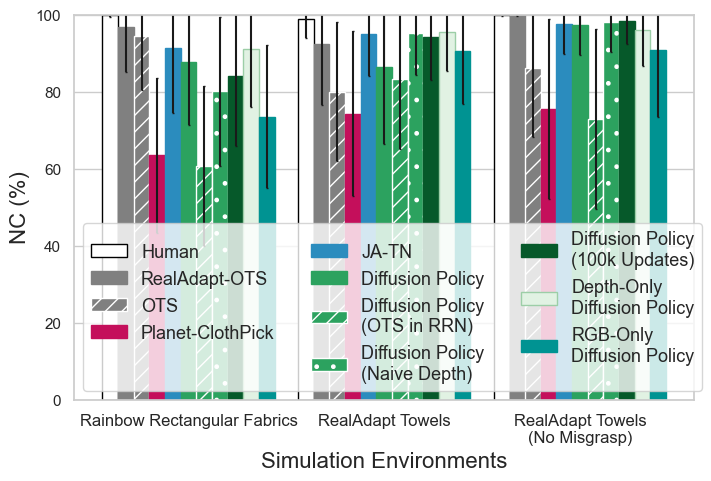}}}
    \subfloat[Normalised Improvement]{{\includegraphics[width=0.33\textwidth]{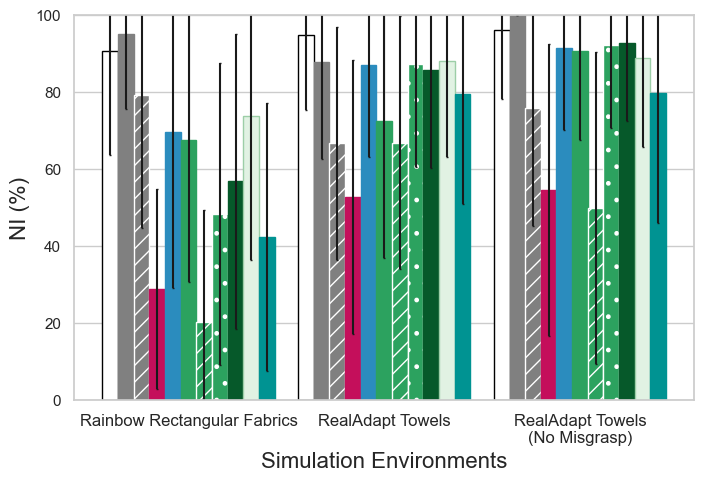}}}
    \subfloat[Success Rate]{{\includegraphics[width=0.33\textwidth]{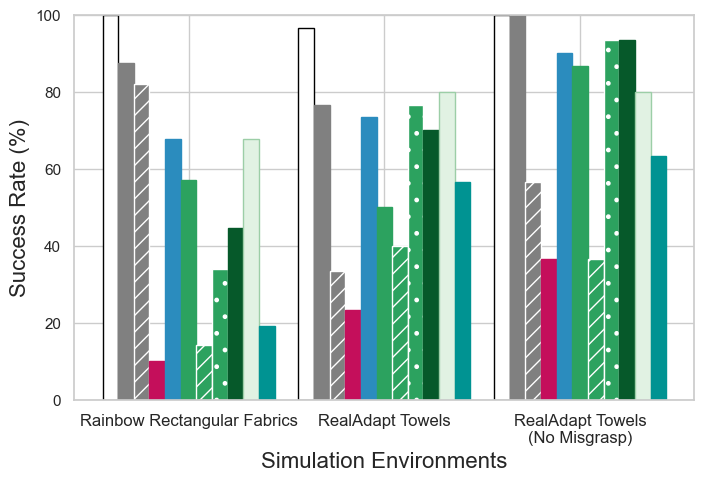}}}
    \caption{DRAPER Ablation Study in Simulation Environments with Various Controllers. A trajectory is regarded as a failure if a policy cannot flatten the fabrics and achieve a normalised coverage (NC) above 99\% within 30 steps. By default, all neural controllers are trained using DRAPER's training strategy. Note that each colour bar represents a single trained policy examined in these three simulation environments. The newly proposed flattening Oracle policy RealAdapt-OTS consistently outperforms the old Oracle Towel Smoothing (OTS). Neural controllers trained with RealAdapt-OTS in RealAdapt Towels generalise better than those trained with OTS in \textit{Rainbow Rectangular Fabrics} \textbf{(OTS in RRN)}. Although naive depth processing \textbf{(Naive Depth)} can help the neural controller improve its performance on the benchmark environment it trains on, it generalises poorly compared to the one trained with DRAPER's setting. Among the neural controllers, JA-TN shows the best generalisability across the tested simulation benchmarks. Depth-Only diffusion policy and JA-TN demonstrate the best generalisability among the neural controllers. }
    
    \label{fig:abaltion-sim}
\end{figure*}

\subsection{Generalisability of DRAPER}

To demonstrate generalisability of our deployment framework DRAPER, we conduct experiments with 4 different neural controllers and 6 different fabrics on 4 different shaping tasks. We also examine the neural controllers on all-corner inward folding (4 steps), corners-edge inward folding (4 steps) and diagonal-cross foldings (2 steps) with extra 2 steps to allow controllers to recover from mistakes. As a control, we also provide a human baselines with a robot executing actions entered via a pick-and-place human interface.

We train PlaNet-ClothPick, JA-TN and diffusion policies (50, 80 and 100 thousand updates respectively) with RGB as input and DRAPER's training protocol. Note that PlaNet-ClothPick is trained on a dataset with 20 thousands transitional steps (instead of its original 1 million) using its specially engineered collection policy \cite{kadi2024planet} and the latter two policies with 2000 successful trajectories created by our new real-world adapted Oracle policy, RealAdapt-OTS, so that all policies will have a similar amount of training data.

Evaluation for flattening controllers in physical trials typically employs step-wise normalised coverage, normalised improvement (requiring real-time cloth segmentation), as well as operational efficiency \cite{hoque2022visuospatial, lin2022learning}. Table I shows that all neural policies can transfer successfully to real-world scenarios for Fabrics C and D, but all struggle with flattening Fabric A. JA-TN performs the best in real-world in general. Compared to human performance, all methods still need improvement especially with soft fabrics like Fabric A. 

On the other hand, assessing folding tasks is more complex due to task variety and the self-occlusive nature of goal states, often necessitating human judgement. IoU serves as a reasonable proxy metric for step-wise folding performance in real world \cite{weng2022fabricflownet, mo2022foldsformer, lee2021learning}. The initial state of the fabric are required to be wholly inside the camera view. We only test on the fabrics with low stiffness, such as Fabric A and B, so they will not flip back to their last state. We train JA-TN and diffusion policies with 1000 oracle-based demonstrations \cite{kadi2024mjtn} in in our newly proposed simulation environment \textit{RealAdapt Towels}. We also adopt the off-the-shelf Foldsformer model with its original deployment strategy as the baseline for folding comparison. 

Table \ref{tab:folding} demonstrates that both JA-TN and diffusion folding policy can be deployed successfully in the real world with our DRAPER framework. Diffusion policy shows near human performance whilst JA-TN and Foldsformer demonstrate similar performance to each other yet slightly inferior to the diffusion policy. Despite performing well in all-corner-inward and diagonal-cross foldings, all methods mainly struggle with more complex corners-edge inward folding.

Tables \ref{tab:flattening} and \ref{tab:folding} empirically demonstrate that our DRAPER framework results in robust deployment of various neural controller to achieve towel-shaping tasks in the real world. It also shows room for improvement on flattening soft fabrics (Fabric A) and performing corners-edge inward folding tasks.


\begin{table}[t!]
\centering
\caption{Grasping Analysis by Cloth and Fold Type. \textbf{Miss} represents misgrasping, \textbf{MLT} represents multi-layer grasping, \textbf{CA} represents cloth-attachment, and \textbf{CD} represents cloth-dropping. The use of our custom-built tweezers significantly reduces grasping errors.}
\resizebox{0.47\textwidth}{!}{
\begin{tabular}{c|c|c|c|c|c|c}
\hline
\textbf{Fabric}     & \textbf{Gripper} & \textbf{SR (\%)} & \textbf{Miss (\%)} & \textbf{MTL (\%)} & \textbf{CA (\%)} & \textbf{CD (\%)} \\ \hline
\multirow{2}{*}{E}  & Franka      & 77.78     & 22.22       & 66.67           & 0.00                & 0.00           \\
                    & Tweezers    & 90.00     & 0.00        & 66.67           & 0.00                & 0.00           \\ \hline
\multirow{2}{*}{C}  & Franka      & 80.00     & 20.00       & 55.56           & 5.56                & 5.56           \\
                    & Tweezers    & 93.75     & 0.00        & 61.11           & 0.00                & 0.00           \\ \hline
\multirow{2}{*}{B}  & Franka      & 83.33     & 16.67       & 69.44           & 2.78                & 5.56           \\
                    & Tweezers    & 92.00     & 8.00        & 59.38           & 0.00                & 0.00           \\ \hline
\multirow{2}{*}{2-Fold} & Franka      & 80.56     & 19.44       & 55.56           & 5.56                & 5.56           \\
                    & Tweezers    & 91.67     & 8.33        & 61.11           & 0.00                & 0.00           \\ \hline
\multirow{2}{*}{3-Fold} & Franka      & 77.78     & 22.22       & 69.44           & 2.78                & 5.56           \\
                    & Tweezers    & 100.00    & 0.00        & 59.38           & 0.00                & 0.00           \\ \hline
\end{tabular}
}
\label{tab:cloth_fold_analysis}
\end{table}

\subsection{Ablation Study of DRAPER}

In this ablation study, we aim to study the effect of different components of the DRAPER framework. We conduct this investigation in both simulation and real-robot setups. Unlike the previous section, we train all diffusion policies with 50,000 updates to clearly demonstrate the effect of the framework.

For the simulation evaluation, as shown in Figure \ref{fig:abaltion-sim}, we create a new testing environment, \textit{RealAdapt Towels (No Misgrasp)}, where we disable the misgrasping scenarios, so that the actual ability of these policies can be demonstrated accurately. Additionally, the real-world experiments, as shown in Table \ref{tab:ablation-real}, indicate that all components of DRAPER are critical for reliable deployment of the neural controllers. Also, training purely on depth inputs and doubling the training steps of RGB-D diffusion policy can help improve performance across different simulation settings.

\subsection{Grasping with Tweezers-extended Gripper}

We evaluate the tweezers-extended gripper against the default Franka gripper aiming to minimise misgrasping, multilayer grasping, cloth dropping and cloth attachment. We  encompass a diverse range of cloth conditions including three cloth types (silk, cotton, and polyester), various cloth states (Flat, 2-folds, 3-folds), and multiple grasping scenarios (4 corners, 4 edges, and 2 random cloth points). Table \ref{tab:cloth_fold_analysis} indicates that the tweezers perform better compared to the Franka gripper across all pick types, particularly excelling in complex graspings on four edges and corners with over 90\% success rate. Misgrasp rates for the tweezers remain consistently below 10\%, while the Franka Gripper shows higher rates between 16-22\%. Both grippers perform similarly for multilayer handling, but the tweezers completely avoid cloth drops and attachment issues, which occur more frequently with the default gripper. Furthermore, the tweezers maintain consistent performance across all cloth types, performing particularly well with delicate fabrics like silk. 

\section{Discussion}
Many neural controllers in towel shaping have only been tested with cloths of particular colour, thickness, and initial positions in camera view \cite{mo2022foldsformer, kadi2024mjtn, kadi2024planet}; their generalisability to other fabric contexts and robot platform is uncertain. Our DRAPER framework allows us to deploy four neural policies, including a diffusion policy, on two different robotic platforms:  Franka Emika Panda with an "eye-in-hand" setting and UR3e with a "hand-to-eye" setting. 

Through our experiments, we find that soft fabrics with high internal friction pose challenges for flattening methods due to the difficulty in separating different parts of the fabric. Conversely, small and stiff fabrics are problematic for folding with a single end-effector as they tend to flip back easily. 

\paragraph{Limitation} Although we successfully deploy numerous simulation-trained neural controllers in real-world setups, our current robotic systems, i.e., the tested neural controllers with DRAPER real-world deployment, faces several limitations: (1) it struggles to automatically detect success states accurately; (2) it occasionally produces helical and spiral path  trajectories; (3) it still experiences minor misgrasping issues; and, (4) for all folding controllers, the initial state of the folding tasks required to be completely inside the field of vision. 



\paragraph{Future} We plan to apply the proposed method to more difficult tasks, such as flattening and folding different garment types, such as T-shirts and trousers. We want to further investigate the capacity of diffusion policy through making folding agnostic to the cloth's initial state and achieve \textit{folding-from-crumpled} in the real world. Integrating depth estimation along with a tactile sensor may help to improve the dexterity of the grasping and substantially reduce the effect of grasping multiple layers of the articles, which may be vital for the successful manipulation of complex garment manipulation. 

\section{Conclusion}
We introduce a set of robust robotic deployment techniques, collectively known as DRAPER deployment framework, that enables effective real-world deployment of pick-and-place cloth-shaping neural controllers. We perform the first comprehensive comparison of four SotA neural controllers in real-world settings, evaluating shaping of towels with varying size, colour, shape, material and thickness. This includes the first application of a visuomotor diffusion policy on this task. Our method also demonstrates the versatility of the proposed protocol across different robotic platforms. Ultimately, this research contributes valuable insights and practical solutions to robotic cloth manipulation, paving the way for more advanced and reliable systems in real-world cloth manipulation.

\section{Acknowledgement}
We wish to thank Mr Nicol Thomson at the University of St Andrews for setting up the UR3e robot as well as printing the base and integrating the tweezers into the gripper.


\bibliographystyle{IEEEtran}
\bibliography{IEEEabrv,ref}

\newpage

\end{document}